\documentclass[conference,a4paper]{APSIPA2021}
\usepackage{amsmath}
\usepackage{graphicx}
\usepackage{multirow}
\usepackage{threeparttable}
\usepackage[backend=biber,style=ieee,]{biblatex}
\graphicspath{{figs/}}
\usepackage{booktabs}
\usepackage{etoolbox}
\usepackage{siunitx}
\usepackage{algorithm,algpseudocode}
\usepackage[caption=false, font=footnotesize]{subfig} 
\usepackage{url}

\usepackage{geometry}
\newcommand*\rot{\rotatebox{90}}

\usepackage{xspace}
\makeatletter
\DeclareRobustCommand\onedot{\futurelet\@let@token\@onedot}
\def\@onedot{\ifx\@let@token.\else.\null\fi\xspace}

\def\ie{{\it i.e}\onedot}

\def\etal{{\it et al}\onedot}
\makeatother

\begin{document}

\title{Hindering Adversarial Attacks with Multiple Encrypted Patch Embeddings}

\author{AprilPyone MaungMaung$^1$, Isao Echizen$^{1,2}$, and Hitoshi Kiya$^{3}$ \\
\small{$^1$National Institute of Informatics, Tokyo, Japan\qquad $^2$University of Tokyo, Tokyo, Japan} \qquad
\small{$^3$Tokyo Metropolitan University, Tokyo, Japan}
}

\maketitle

\begin{abstract}
In this paper, we propose a new key-based defense focusing on both efficiency and robustness. Although the previous key-based defense seems effective in defending against adversarial examples, carefully designed adaptive attacks can bypass the previous defense, and it is difficult to train the previous defense on large datasets like ImageNet.
We build upon the previous defense with two major improvements: (1) efficient training and (2) optional randomization.
The proposed defense utilizes one or more secret patch embeddings and classifier heads with a pre-trained isotropic network. When more than one secret embeddings are used, the proposed defense enables randomization on inference. Experiments were carried out on the ImageNet dataset, and the proposed defense was evaluated against an arsenal of state-of-the-art attacks, including adaptive ones. The results show that the proposed defense achieves a high robust accuracy and a comparable clean accuracy compared to the previous key-based defense. 
\end{abstract}

\section{Introduction}
Deep learning has brought major breakthroughs in many applications~\cite{lecun2015deep}.
Some notable examples are visual recognition~\cite{he2016deep}, natural language processing~\cite{vaswani2017attention}, and speech recognition~\cite{graves2013speech}.
Despite the remarkable performance, machine learning (ML) in general, including deep learning, is vulnerable to a wide range of attacks.
Particularly, many ML algorithms, including deep neural networks, are sensitive to carefully perturbed data points known as adversarial examples that are imperceptible to humans, but they cause ML models to make erroneous predictions with high confidence~\cite{szegedy2013intriguing,biggio2013evasion,goodfellow2015explaining}.
Not only in the digital domain, adversarial examples can also appear in the physical domain.
One work shows that an adversarial example can be photographed with a smartphone, and the taken picture can still fool a neural network~\cite{kurakin2018adversarial}.
Similarly, another work demonstrated that it is possible to construct 3D adversarial objects~\cite{athalye2018synthesizing}.
Such physical adversarial attacks may cause serious and dangerous problems by applying paintings or stickers on physical objects~\cite{papernot2017practical,eykholt2018robust}.
For example, self-driving cars may misclassify a ``Stop'' sign as a ``Speed Limit''~\cite{eykholt2018robust}, and face recognition systems may authenticate unauthorized users as authorized ones~\cite{xu2020adversarial,komkov2021advhat,sharif2016accessorize,guetta2021dodging}.

As adversarial examples are an obvious threat, numerous methods for generating adversarial examples (attack methods) and defenses against them have been proposed in the literature~\cite{yuan2019adversarial}.
Adversarial attacks and defenses have entered into an arms race, and attack methods are ahead of defense ones.
Once defense mechanisms are known to the attacker, adaptive attacks can be carried out~\cite{athalye2018obfuscated,tramer2020adaptive}.
Most of the defense methods either reduce the classification or are completely broken.
Therefore, defending against adversarial examples is still challenging and remains an open problem.

Inspired by cryptography, a new line of research on adversarial defense has focused the use of secret keys so that defenders have some information advantage over attackers~\cite{taran2018bridging,aprilpyone2021block,kiya2022overview,maung2020encryption,iijima2023enhanced}.
Key-based defenses follow Kerckhoffs's second cryptographic principle, which states that a system should not require secrecy even if it is exposed to attackers, but the key should be secret~\cite{kerckhoffs1883cryptographic}.
The main idea of the key-based defense is to embed a secret key into the model structure with minimal impact on model performance.
Assuming the key stays secret, an attacker will not obtain any useful information on the model, which will render adversarial attacks ineffective.
The idea of making adversarial attacks expensive or ideally intractable is further supported on a theoretical basis that adversarially robust machine learning could leverage computational hardness, as in cryptography~\cite{garg2020adversarially}.

However, a recent study shows that applying cryptographic principles alone in algorithm design is insufficient to defend against efficient adversarial attacks~\cite{rusu2022hindering}.
To further harden key-based defenses, researchers have also proposed to use implicit neural representation~\cite{rusu2022hindering} and ensembles of key-based defenses~\cite{taran2020machine,maungmaung2021ensemble}.
Nevertheless, although these defenses seem promising, they have yet to tested on large datasets like ImageNet.
Therefore, in this paper, we build upon the idea of key-based defense and propose a novel defense that is not only efficient to hinder adversarial attacks but also scalable to large datasets.
The proposed defense utilizes multiple secret patch embeddings and classifier heads that are controlled by secret keys and a pre-trained isotropic network (with the same depth and resolution across different layers in the network) backbone such as a vision transformer (ViT)~\cite{dosovitskiy2020image} and ConvMixer~\cite{trockman2022patches}.
Our contributioins in this paper are as follows.
\begin{itemize}
  \item We propose a novel defense idea that uses multiple secret patch embeddings.
  \item We show that the proposed defense is computationally feasible even on ImageNet and report empirical results on the strongest attacks.
\end{itemize}
In experiments, the proposed defense is confirmed not only to be efficient against attacks, but also to maintain a comparable clean accuracy.

\section{Related Work}
\subsection{Adversarial Robustness}
There are two distinct strategies in designing adversarial defenses: (1) classifiers are designed in such a way that they are robust against all adversarial examples in a specific adversarial space either empirically (\ie, adversarial training) or in a certified way (\ie, certified defenses), and (2) input data to classifiers are pre-processed in such a way that adversarial examples are ineffective (\ie, input transformation defenses, key-based defenses).

\vspace{2mm}\noindent{\bf Adversarial Training.}
Current empirically robust classifiers utilize adversarial training, which includes adversarial examples in a training set.
Madry \etal approach adversarial training as a robust optimization problem and utilizes projected gradient descent (PGD) adversary under $\ell_\infty$-norm to approximate the worst inputs possible (\ie, adversarial examples)~\cite{madry2018towards}.
As PGD is iterative, the cost of computing PGD-based adversarial examples is expensive.
Much progress has been made to reduce the computation cost of adversarial training as in free adversarial training~\cite{shafahi2019adversarial}, fast adversarial training~\cite{wong2020fast}, and single-step adversarial training~\cite{de2022make}.
However, adversarially trained models (with $\ell_\infty$ norm-bounded perturbation) can still be attacked by $\ell_1$ norm-bounded adversarial examples~\cite{sharma2017attacking}.

\vspace{2mm}\noindent{\bf Certified Defenses.}
Robust classifiers in this direction use formal verification methods in such a way that no adversarial examples exist within some bounds~\cite{wong2018provable,raghunathan2018certified,cohen2019certified,hein2017formal}.
Ideally, these defenses are preferred for achieving certain guarantees..
Although certified defenses are attractive, they can be bypassed by generative perturbation~\cite{poursaeed2018generative} or parametric perturbation (outside of pixel norm ball)~\cite{liu2018beyond}.

\vspace{2mm}\noindent{\bf Input Transformation.}
Adversarial defenses in this strategy aim to find a defensive transform to reduce the impact of adversarial noise or make adversarial attacks ineffective (\ie, computing adversarial noise is either expensive or intractable).
The works in this direction use various transformation methods, such as thermometer encoding~\cite{buckman2018thermometer}, diverse image processing techniques~\cite{guo2018countering,xie2018mitigating}, denoising strategies~\cite{liao2018defense,niu2020limitations}, GAN-based transformation~\cite{song2018pixeldefend}, and so on.\
Although these input transformation-based defenses provided high accuracy at first, they can be attacked by adaptive attacks such as~\cite{athalye2018obfuscated,tramer2020adaptive}.

\vspace{2mm}\noindent{\bf Key-Based Defenses.}
The defenses with a secret key are similar to the input transformation-based defenses.
However, the major difference is that defenders have an information advantage over attackers (\ie, a secret key).
The success of adaptive attacks is based on the ability to approximate a defended model's outputs on arbitrary inputs.
On this insight, key-based defenses hide a model’s decision from attackers by means of training the model with encrypted images.
Such key-based defenses include~\cite{taran2018bridging,aprilpyone2021block}.
However, a recent study pointed out that such key-based defenses are still vulnerable to adaptive attacks~\cite{rusu2022hindering}.
Instead, they proposed a new key-based defense by using neural implicit representation~\cite{rusu2022hindering}.

\subsection{Patch Embedding}
Inspired by the success of the vision transformer (ViT)~\cite{dosovitskiy2020image}, a series of works such as~\cite{trockman2022patches,tolstikhin2021mlp,touvron2022resmlp,chen2021cyclemlp,liu2021pay,hou2022vision} utilizes an isotropic architecture that uses patch embeddings for the first layer.
Especially a recent network architecture, ConvMixer~\cite{trockman2022patches} showed some evidence that patch embeddings play an important role in achieving high performance accuracy.
With hindsight, we fine-tune the patch embedding layer by using block-wise encrypted images to obtain encrypted patch embeddings in this paper.

\subsection{Block-Wise Image Transformation\label{sec:transformation}}
\begin{figure}[t]
\centering
\subfloat[]{\includegraphics[width=0.33\linewidth]{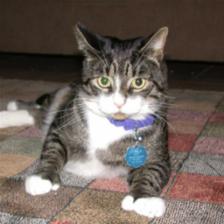}%
\label{fig:cat}}
\hfil
\subfloat[]{\includegraphics[width=0.33\linewidth]{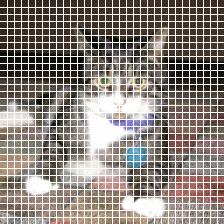}%
\label{fig:cat-grid}}
\hfil
\subfloat[]{\includegraphics[width=0.33\linewidth]{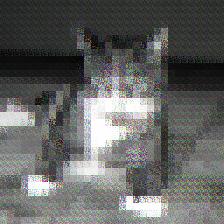}%
\label{fig:cat-shuffle}}
\caption{Block-wise image transformation with block size $M = 7$.\ (a) Plain.\ (b) Block division.\ (c) Block-wise shuffling.\label{fig:block-wise}}
\end{figure}

The previous key-based defense proposed three block-wise transformation with a secret key for an adversarial defense: pixel shuffling, bit flipping, and FFX-based transformation~\cite{aprilpyone2021block}.
In this paper, we adopt block-wise pixel shuffling from~\cite{aprilpyone2021block} for the proposed defense, and the detailed procedure for block-wise pixel shuffling is described as follows.
\begin{enumerate}
  \item Divide a three-channel (RGB) color image, $x$ with $w \times h$ into non-overlapping blocks each with $M \times M$ such that $\{B_1, \ldots, B_i, \ldots, B_{(\frac{h}{M} \times \frac{w}{M})}\}$.
    \item Generate a random permutation vector, $v$ with key $K$, such that \\ $(v_1, \ldots, v_k, \ldots, v_{k'}, \ldots, v_{3M^2})$, where $v_k \neq v_{k'}$ if $k \neq k'$.
    \item For each block $B_i$, \\
      flatten three-channel block of pixels into a vector, $b_i$ such that $b_i = (b_i(1), \ldots, b_i({3M^2}))$,\\
      permute pixels in $b_i$ with $v$ such that
      \begin{equation}
        b'_i(k) = b_i(v_k), k \in \{1, \ldots, 3M^2\},
      \end{equation}
      and reshape the permuted  vector $b'_i$ back into the three-channel block $B'_i$.
    \item Integrate all permuted blocks, $B'_1$ to $B'_{(\frac{h}{M} \times \frac{w}{M})}$ to form a three-channel pixel shuffled image, $x'$.
\end{enumerate}
As an example, Fig.~\ref{fig:block-wise} shows a plain image with visualization of blok division and block-wise shuffling.

\section{Proposed Defense with Encrypted Patch Embeddings}

\subsection{Overview}
\begin{figure}[!t]
\centering
\includegraphics[width=\linewidth]{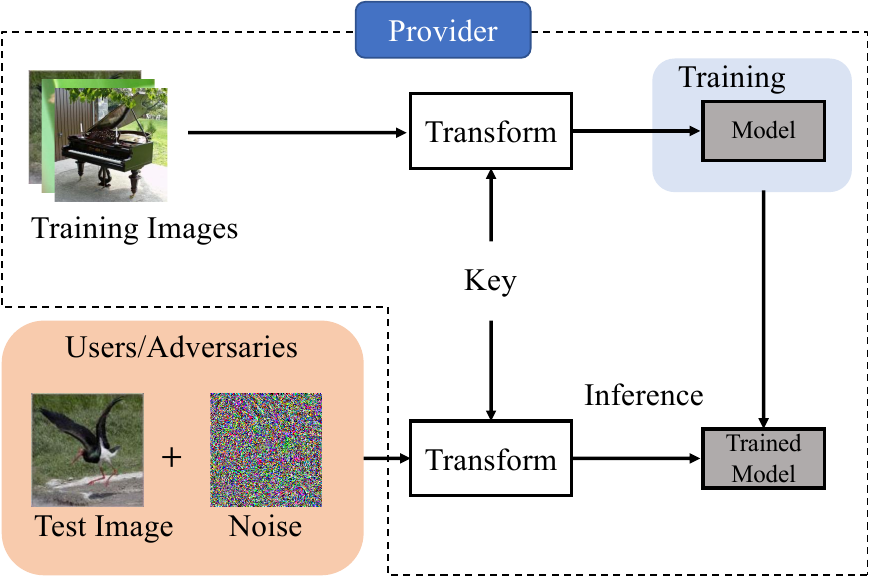}
\caption{Overview of image classification scenario with key-based defense.\label{fig:scenario}}
\end{figure}

We consider an image classification scenario where a provider deploys a key-based defense and provides application programming interface (API) to users (potentially adversarial) for inference as shown in Fig.~\ref{fig:scenario}.

Formally, given a dataset $\mathcal{D}$ with pairs of examples (images and corresponding labels), $\{(x, y) \;|\; x \in \mathcal{X},\; y \in \mathcal{Y}\}$,
a key-based defense in~\cite{aprilpyone2021block} first maps the input space $\mathcal{X}$ to an encrypted space $\mathcal{H}$ by using a blok-wise transformation with a secret key (\ie, $x \mapsto t(x, K)$).
Then, a robust classifier $f_\theta$ parameterized by $\theta$ ($f_\theta: \mathcal{X} \rightarrow \mathcal{Y}$) is trained by using encrypted images $t(x, K) \in \mathcal{H}$.

\begin{figure}[!t]
\centering
\subfloat[Previous~\cite{aprilpyone2021block}]{\includegraphics[width=\linewidth]{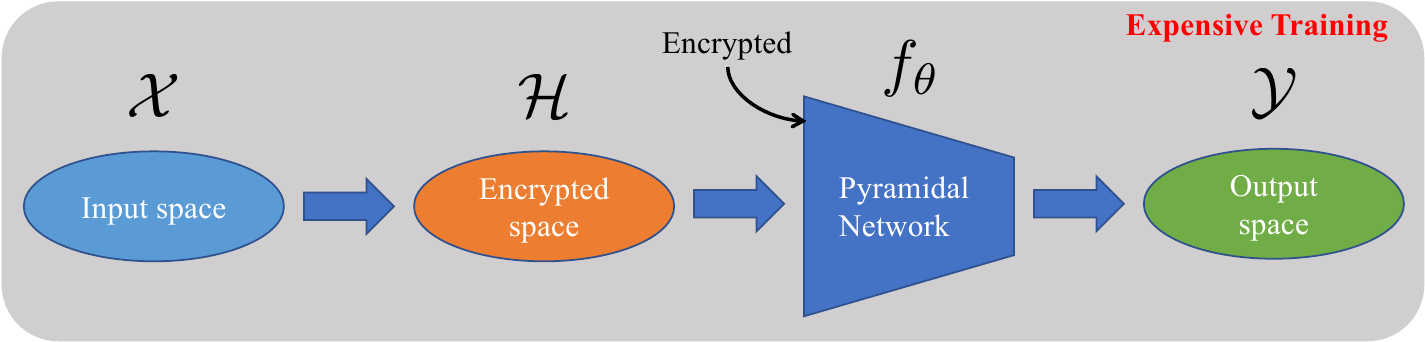}%
\label{fig:previous}}
\hfil
\subfloat[Proposed]{\includegraphics[width=\linewidth]{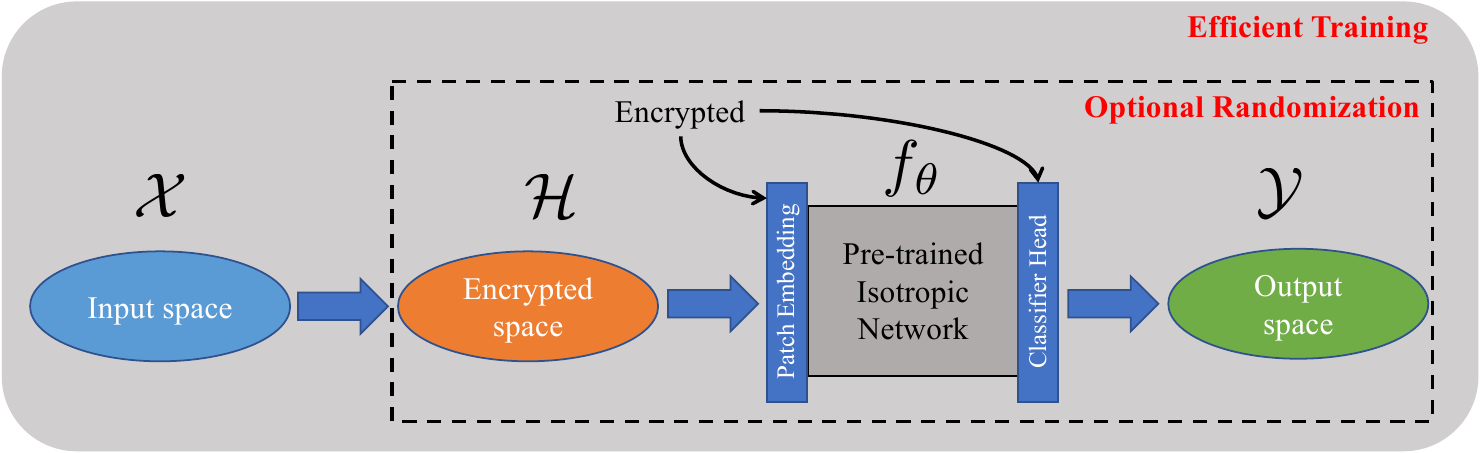}%
\label{fig:proposed}}
\caption{Difference between previous and proposed key-based defense.\label{fig:overview}}
\end{figure}

We build upon the robust classifier $f_\theta$ from~\cite{aprilpyone2021block} with two major improvements: (1) efficient training and (2) optional randomization.\
Figure~\ref{fig:overview} highlights the difference between the previous key-based defense and the proposed one.
The proposed defense leverages pre-trained isotropic networks (such as ViT, ConvMixer, etc.), and fine-tunes only patch embeddings and classifier heads (optionally multiple times with different secret keys).
When more than one secret patch embedding is used, the proposed defense becomes a radomized defense.
The details of training and inference pipelines for the proposed defense are described in the following subsections.

\begin{figure}[!t]
\centering
\subfloat[Training]{\includegraphics[width=\linewidth]{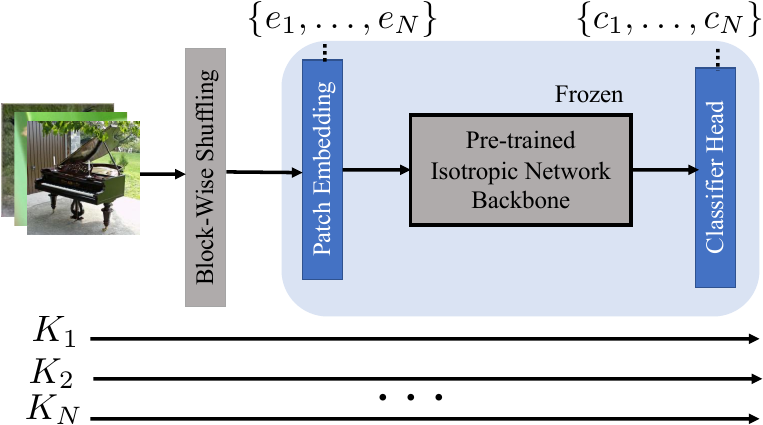}%
\label{fig:train}}
\hfil
\subfloat[Inference]{\includegraphics[width=\linewidth]{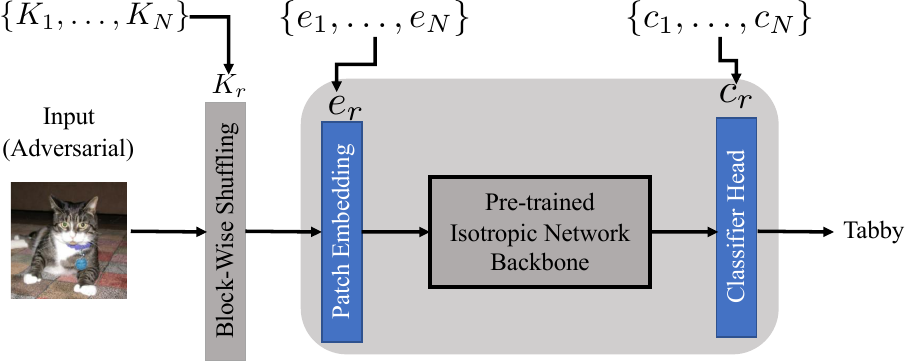}%
\label{fig:test}}
\caption{Training and inference pipelines for proposed defense. During training, pre-trained isotropic network backbone is frozen.\label{fig:pipeline}}
\end{figure}

\subsection{Training}
The proposed defense utilizes a patch-based pre-trained isotropic network (such as ViT, ConvMixer, etc.).
The training pipeline of the proposed defense is depicted in Fig.~\ref{fig:train}.
To obtain encrypted patch embeddings, $\{e_1, \ldots, e_N\}$ and encrypted classifier heads, $\{c_1, \ldots, c_N\}$, only the patch embedding layer and classifier head in a pre-trained isotropic network is fine-tuned for $N$ times by using encrypted images that are encrypted by a block-wise pixel shuffling $t$ with $N$ secret keys $\{K_1, \ldots, K_N\}$.
The detailed procedure of $t$ was previously described in Section~\ref{sec:transformation}.

\subsection{Inference}
Figure~\ref{fig:test} shows the inference process of the proposed defense.
We first sample a random key $K_r$ from a pool of keys, $\{K_1, \ldots, K_N\}$, and select the corresponding patch embedding, $e_r \in \{e_1, \ldots, e_N\}$, and classifier head, $\{c_1, \ldots, c_N\}$.
During inference, a test image (potentially adversarial) is transformed with $K_r$.
Then, the corresponding patch embedding $e_r$ and classifier head $c_r$ are selected and attached to the pre-trained isotropic network backbone to carry out the forward pass.
Note that when $N$ is greater than $1$, the proposed defense provides randomization ability for each inference.

\subsection{Threat Model}
We focus on hindering adversarial attacks on a classifier $f_\theta$, which takes block-wise transformed input, $t(x, K)$ with a secret key $K$.
To evaluate the proposed defense, we assume a gray-box threat model, where the attacker has full knowledge of the defense mechanism, pre-trained isotropic network backbone weights, and training data, except a secret key, encrypted patch embeddings weights, and classifier heads.
The attacker performs state-of-the-art both targeted and untargeted attacks under $\ell_\infty$ and $\ell_2$ norms.

\section{Experiments and Discussion}
\subsection{Setup}
We carried out experiments for the proposed defense on ImageNet-1k classification.
The ImageNet-1k dataset (with 1000 classes) consists of 1.28 million color images for training and 50,000 color images for validation~\cite{ILSVRC15}.
To implement the proposed defense, we utilized a publicly available pre-trained isotropic network, ConvMixer-768/32\footnote{\scriptsize \url{https://github.com/locuslab/convmixer}} model, which uses a patch size value of 7 on $224 \times 224$ images.
To inline with the patch size of the patch embedding in the pre-trained model, we also set the same value of 7 for the block size value in the key-based transformation (\ie, $M = 7$).
We fine-tuned only the patch embedding layer and classifier head with the same learning rate value of 0.01 as in the original training settings without default augmentation for only one epoch.
The fine-tuning process was repeated $N$ times to produce $N$ secret patch embeddings and classifier heads with $N$ secret keys.
In our experiments, we used $N = 5$.

For evaluation, we deployed AutoAttack (AA) strategy, which is an ensemble of strong, diverse attacks~\cite{croce2020reliable}.
We used AA's ``standard'' version with a perturbation budget value of $4/225$ under $\ell_\infty$ norm and a value of $0.5$ under $\ell_2$ norm for all attacks.
To account for adaptive attacks, we also utilized a ``random'' version, which implements Expectation over Transformation (EoT) as described in~\cite{athalye2018obfuscated}.

\robustify\bfseries
\sisetup{table-parse-only,detect-weight=true,round-mode=places,round-precision=2}
\begin{table*}[tbp]
  \setlength{\extrarowheight}{.2em}
  \caption{Clean and robust accuracy (\SI{}{\percent}) of proposed defense and state-of-the-art classifiers\label{tab:res-standard}}
  \centering
  \resizebox{\textwidth}{!}{%
  \begin{tabular}{clSSSSScc}
  \toprule
  & & & \multicolumn{2}{c}{$\ell_\infty$} & \multicolumn{2}{c}{$\ell_2$} &\\
  \cmidrule{4-7}
  &  & {Clean} & {Robust (Scaneraio 1)} & {Robust (Scenario 2)} & {Robust (Scenario 1)} & {Robust (Scenario 2)} & \\
  & {Model} & {Accuracy} & {Accuracy} & {Accuracy} & {Accuracy} & {Accuracy} & {Defense}\\
  \midrule
    \multirow{3}{*}{\rot{Proposed}} & ConvMixer-768/32 ($N=1$) & 71.9820 & 64.7380 & \bfseries \num{70.6480} & \bfseries \num{70.1640} & \bfseries \num{71.4680} & \multirow{3}{*}{Key}\\
                                    & ConvMixer-768/32 ($N=3$) & 71.4340 & 64.7480 & 70.0500 & 69.6380 & 70.8560 & \\
                                    & ConvMixer-768/32 ($N=5$) & 71.5500 & 64.7400 & 70.0540 & 69.7240 & 70.9180 & \\
  \midrule
                                    & ConvMixer-768/32 & \bfseries \num{80.154} & 0.0 & {--} & 0.3180 & {--} & {Plain}\\
                                    & $^{\dagger}$ResNet50 (\cite{aprilpyone2021block}) & 75.6900 & \bfseries \num{66.9480} & 10.062  & {--} & {--} & {Key}\\
  \midrule
                                    & Swin-L (\cite{liu2023comprehensive}) & 78.92 & 59.56 & {--} & {--} & {--} & \multirow{3}{*}{AT}\\
                                    & ConvNeXt-L (\cite{liu2023comprehensive}) & 78.02 & 58.48 & {--} & {--} & {--} & \\
                                    & ConvNeXt-L + ConvStem (\cite{singh2023revisiting}) & 77.00 & 57.70 & {--} & {--} & {--} & \\
  \bottomrule

  \multicolumn{3}{l}{$^{\dagger}$ The model was tested according to its stated threat model.}
  \end{tabular}
}
\end{table*}

\robustify\bfseries
\sisetup{table-parse-only,detect-weight=true,round-mode=places,round-precision=2}
\begin{table}[tbp]
  \caption{Attack success rate (\SI{}{\percent}) of adaptive attacks on proposed defense\label{tab:res-adaptive}}
  \centering
  \begin{tabular}{lSS}
  \toprule
  {EoT} & {ASR (AA --- $\ell_\infty$)} & {ASR (AA --- $\ell_2$)}\\
  \midrule
  {EoT ($N = 1$)} & 8.4577 & 7.1287\\
  {EoT ($N = 3$)} & 10.9252 & 7.6162\\
  {EoT ($N = 5$)} & 14.7716 & 8.4075\\
  \bottomrule
  \end{tabular}
\end{table}

\subsection{Results}
We performed ImageNet-1k classification of both clean images and adversarial examples on the proposed defended classifier.

\vspace{2mm}\noindent{\bf Transfer Attacks.}
We generated adversarial examples for two scenarios.
\begin{itemize}
  \item {\bf Scenario 1.} Since the proposed defense is built on top of the pre-trained model, it is natural to generate adversarial examples directly on the pre-trained model (which includes pre-trained patch embeddings, isotropic network backbone, and a classifier head).
  \item {\bf Scenario 2.} We assume the attacker has knowledge of the proposed defense mechanism. The attacker may fine-tune the pre-trained model with a guessed key. Thus, we also simulated this scenario to evaluate the proposed defense.
\end{itemize}

We carried out attacks on both scenarios, and the results are shown in Table~\ref{tab:res-standard}, where (--) denotes ``there is no reported results'' or ``not applicable''.
Both clean and robust accuracies were calculated on the validation set of ImageNet (50,000) images.
Although the un-defended model (plain) achieved the highest accuracy, it was most vulnerable to all attacks.
The results suggest that adversarial examples generated on the pre-trained model were not effective on the proposed defense method with secret patch embeddings and a classifier head.
Under attacks, the proposed method achieved comparable accuracy for Scenario 1 and higher accuracy for Scenario 2.
However, the proposed method reduced clean accuracy, and we shall improve it in our future work.

\vspace{2mm}\noindent{\bf Adaptive Attacks.}
When using $N > 1$, the proposed utilizes random secret patch embeddings and a classifier head from $N$ options.
Therefore, the proposed defense produces a randomized prediction on each inference.
To counter the proposed randomized defense, we assume that the attacker prepares multiple secret patch embeddings as in the proposed defense.
We carried out EoT attacks on the attacker's version of the proposed defense with $N \ge 1$ and evaluated the proposed defense.
Table~\ref{tab:res-adaptive} presents the results of such adaptive attacks.
For this experiment, attack success rate (ASR) was calculated for randomly selected 1000 images that were correctly classified by the proposed defense model with $N = 5$.
Although the EoT attack was improved when using $N = 5$, the ASR was still low.

\subsection{Discussion}

\noindent{\bf Comparison with state-of-the-art methods.}
For experiments, we implemented the proposed method in a relatively small isotropic model, ConvMixer-768/32, which contains about $21.1$ million parameters (\ie, even smaller than ResNet50).
We compare the proposed defense with the previous key-based defense and the top 3 adversarially trained (AT) models~\cite{croce2020robustbench} in terms of clean and robust accuracy.
Note that the key-based defense, including the proposed one, has an information advantage (secret key) over attackers.
In contrast, AT models are trained to be robust against all adversarial examples under a specific adversarial space.
Overall, the proposed defense improved the robust accuracy but reduced the clean accuracy.

\vspace{2mm}\noindent{\bf Advantages.}
The main advantage of the proposed method is its feasibility.
Much of the adversarial robustness has been done on low-resolution datasets like CIFAR-10~\cite{krizhevsky2009learning}.
In contrast, much less is known for ImageNet due to high computing requirements.
The proposed method overcomes this barrier by effectively re-using pre-trained weights and only fine-tuning a patch embedding layer and a classifier head.

\vspace{2mm}\noindent{\bf Limitations.}
The proposed method shows the potential of making publicly available pre-trained models to be robust against adversarial examples.
However, further evaluation and investigation of the proposed method against adaptive attacks is required.
In addition, the proposed method in its current form reduces classification accuracy on clean images, although robust accuracy is improved.
We shall address these limitations in our future work.

\section{Conclusions}
In this paper, we proposed a novel key-based defense by using one or more secret patch embeddings and classifier heads with a pre-trained isotropic network.
The proposed method is not only efficient in training even for ImageNet but also effective in hindering strong adversarial attacks.
Experiment results showed that the proposed method achieved high robust accuracy against a suit of powerful attacks under different norm metrics compared to the previous key-based defense.
In addition, we also carried out adaptive attacks to further evaluate the effectiveness of the proposed method.
The results also confirmed a low attack success rate suggesting the proposed method was resistant to such adaptive attacks.

\printbibliography

\end{document}